\documentclass[lettersize,journal]{IEEEtran}
\usepackage{amsmath,amsfonts}
\usepackage{algorithmic}
\usepackage{algorithm}
\usepackage{array}
\usepackage[caption=false,font=normalsize,labelfont=sf,textfont=sf]{subfig}
\usepackage{textcomp}
\usepackage{stfloats}
\usepackage{url}
\usepackage{verbatim}
\usepackage{graphicx}
\usepackage{cite}
\usepackage{booktabs}
\usepackage{bm}
\usepackage{multirow}
\newcommand{\eg}{\textit{e.g.}}
\newcommand{\ie}{\textit{i.e.}}

\usepackage{tikz}
\usepackage{comment}
\usepackage{amsmath,amssymb} 
\usepackage{color}
\definecolor{dodgeblue}{RGB}{30,144,255}
\definecolor{lightgreen}{RGB}{0,157,0}

\usepackage{diagbox}

\usepackage[breaklinks=true,letterpaper=true,colorlinks,bookmarks=false]{hyperref}

\usepackage{colortbl}
\definecolor{mygray}{gray}{.9}
\usepackage{array, boldline, makecell, booktabs}

\begin{document}

\title{Solve the Puzzle of Instance Segmentation in Videos: A Weakly Supervised Framework with Spatio-Temporal Collaboration}

\author{Liqi Yan, Qifan Wang, Siqi Ma, Jingang Wang, Changbin Yu*
\thanks{This work is in part supported by the Shandong Provincial Natural Science Fund (2022HWYQ-081) and the National Science Foundation of China Key Project (U21A20488).}
\thanks{L. Yan is enrolled at Westlake Institute for Advanced Study, Fudan University, China (e-mail: lqyan18@fudan.edu.cn)}
\thanks{Q. Wang is with Meta AI, USA (e-mail: wqfcr@fb.com).}
\thanks{S. Ma and L. Yan are also with School of Engineering, Westlake University, China (e-mail: \{yanliqi,masiqi\}@westlake.edu.cn).}
\thanks{J. Wang is with Meituan, China (e-mail: wangjingang02@meituan.com).}
\thanks{C. Yu is with College of Artificial Intelligence and Big Data for Medical Science, Shandong First Medical University \& Shandong Academy of Medical Sciences, China, and adjunct with the Institute for Intelligent Robots, Fudan University, China (corresponding e-mail: Yu\_lab@sdfmu.edu.cn).}
\thanks{* is corresponding author.}
}

\markboth{IEEE Transactions on Circuits and Systems for Video Technology, ~Vol.~X, No.~X, August~2022}%
{Shell \MakeLowercase{\textit{et al.}}: A Sample Article Using IEEEtran.cls for IEEE Journals}


\maketitle

\begin{abstract}
Instance segmentation in videos, which aims to segment and track multiple objects in video frames, has garnered a flurry of research attention in recent years.
In this paper, we present a novel weakly supervised framework with \textbf{S}patio-\textbf{T}emporal \textbf{C}ollaboration for instance \textbf{Seg}mentation in videos, namely \textbf{STC-Seg}. Concretely, STC-Seg demonstrates four contributions. First, we leverage the complementary representations from unsupervised depth estimation and optical flow to produce effective pseudo-labels for training deep networks and predicting high-quality instance masks. Second, to enhance the mask generation, we devise a puzzle loss, which enables end-to-end training using box-level annotations. Third, our tracking module jointly utilizes bounding-box diagonal points with spatio-temporal discrepancy to model movements, which largely improves the robustness to different object appearances. Finally, our framework is flexible and enables image-level instance segmentation methods to operate the video-level task. 
We conduct an extensive set of experiments on the KITTI MOTS and YT-VIS  datasets. Experimental results demonstrate that our method achieves strong performance and even outperforms fully supervised TrackR-CNN and MaskTrack R-CNN. We believe that STC-Seg can be a valuable addition to the community,   as it reflects the tip of an iceberg about the innovative opportunities in the weakly supervised paradigm for instance segmentation in videos.
\end{abstract}


\section{Introduction}
\label{sec:intro}
The importance of the weakly supervised paradigm cannot be overstated, as it permeates through every corner of recent advances in  computer vision~\cite{liu2021densernet,cheng2022physical,liang2022triangulation} to reduce the annotation cost~\cite{cui2021tf,liu2021densernet}. In contrast to object segmentation~\cite{lu2020learning,perazzi2016benchmark,porikli2009compressed,zhao2016real}, for instance segmentation in videos~\cite{liu2021sg,voigtlaender2019mots,yang2019video}, dense annotations need to depict accurate instance boundaries as well as object temporal consistency across frames, which is extremely labor-intensive to build datasets at scale required to train a deep network. Although a large body of works on weakly supervised image instance segmentation have been discussed in literature~\cite{khoreva2017simple,laradji2019masks,zhou2018weakly}, the exploration in video domain remains largely unavailable until fairly recently~\cite{hoyer2021three,lin2021query,liu2021weakly,lu2020learning,wang2019fast}. Therefore, understanding and improving the weakly supervised methods of instance segmentation in videos are the key enablers for future advances of this critical task in computer vision.
\begin{figure*}[t]
    \centering
    \includegraphics[width=0.95\linewidth]{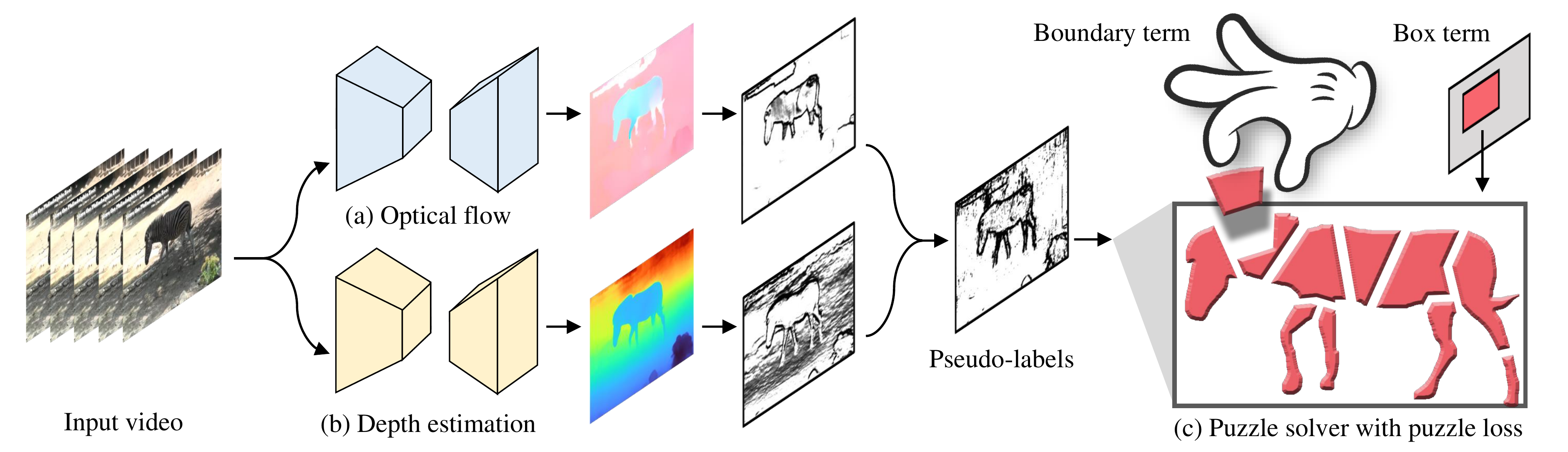}
    \caption{\textbf{Working pipeline of STC-Seg.} The pseudo-labels are generated from spatial and temporal signals, which capture the instance boundary with more accurate edges. 
    Our puzzle solver supervises mask predictions to assemble each sub-region mask together to match the shape of the target with box annotations.
    }
    \label{fig:intro1}
    \vspace{-8pt}
\end{figure*}\\
\indent Developing a weakly supervised framework is a  challenging task. One core objective is to devise reliable \textbf{pseudo-labels} and \textbf{loss function} to perform effective supervision~\cite{khoreva2017simple,wei2018revisiting}. To date, a popular convention is to use class labels produced by Class Activation Map (CAM) or its variants \cite{selvaraju2017grad,zhou2016learning} to supervise image instance segmentation \cite{ahn2019weakly,cholakkal2019object,laradji2019masks,zhou2018weakly,zhu2019learning}. However,  the CAM-based supervision  signal  may  capture  spurious  dependencies  in training due to two daunting issues: 1) It can only identify the most salient features on object regions, which often lose the overall object structures, resulting in partial instance segmentation \cite{ahn2018learning,lee2019ficklenet,yang2019towards}; 2) It cannot separate overlapping instances of the same class and generally lose the capacity to  describe individual targets, when dealing with multiple instances present in an image \cite{ahn2019weakly,wang2020self,cui2022dg}. The challenge is further compounded by instance appearance changes caused by occlusion or truncation \cite{chen2020banet,hur2016joint}. Thus, though CAM is outstanding in semantic segmentation, it does not perform well in instance segmentation. \textit{Under the circumstance, there is a necessity to explore novel weakly supervised approaches with more effective pseudo-labels for video-level instance segmentation}.\\
\indent Aside from the CAM-based family, a line of research has attempted to tackle image instance segmentation with box-level annotations~\cite{dai2015boxsup,hsu2019weakly,kulharia2020box2seg,rajchl2016deepcut}. 
Albeit achieving improvements over CAM-based solutions, they generally have complicated training pipelines, which incur a large computational budget and long supervision schedule. To address this issue, a recent work, BoxInst \cite{tian2021boxinst} introduces a simple yet effective mask loss for training, including a projection term and an affinity term. The first term minimizes the discrepancy between the horizontal and vertical projections of the predicted mask and the ground-truth box. The second term is to identify confident pixel pairs with the same label to explore the instance boundary. With the same supervision level, BoxInst achieves significant improvement over the prior efforts using box annotations \cite{arun2020weakly,papandreou2015weakly,song2019box}. \textit{This successful exploration highlights the importance of loss function to train deep networks in a weakly supervised fashion for the segmentation task.}\\
\indent On the basis of the preceding lessons, one could argue that box-supervised instance segmentation in videos is feasible. \textit{In view of the nature of video data, our conjecture is that one can leverage the rich spatio-temporal information in video to develop reliable pseudo-labels for enhancing the box-level supervision.} In particular, optical flow captures the temporal motion among instances which ensures the same instances have similar flow vectors (Fig. \ref{fig:intro1}a), while depth estimation provides the spatial relation between instance and background (Fig. \ref{fig:intro1}b). We leverage the complementary  representation  of  \textit{\textbf{spatio-temporal  signals}}  to  produce  high-quality pseudo-label to supervise instance segmentation in videos. To enable effective training with the proposed pseudo-labels, we propose a novel \textit{\textbf{puzzle loss}} that organizes learning in a manner compatible with box annotations, including a boundary term and a box term. The two terms collaboratively resolve the puzzle of how to assemble suitable sub-region masks that match the shape of the instance, facilitating the trained model to be boundary sensitive for fine-grained prediction (Fig. \ref{fig:intro1}c).
Furthermore, in contrast to previous efforts \cite{voigtlaender2019mots,yang2019video}, which use simple matching algorithms for tracking, we introduce an \textit{\textbf{enhanced tracking}} module that tracks diagonal points across frames and ensures spatio-temporal consistency for instance movement. To establish the conjecture, our work essentially delivers the following contributions:

\begin{itemize}
    \item We develop a Spatio-Temporal Collaboration framework for instance Segmentation (STC-Seg) in videos, which leverages the complementary representations of depth estimation and optical flow to produce high-quality pseudo-labels for training the deep network. 
    \item We design an effective puzzle loss to assemble mask predictions on each sub-region together in a self-supervised manner. A strong tracking module is implemented with spatio-temporal discrepancy for robust object appearance changes.
    \item The flexibility of our STC-Seg enables weakly supervised instance segmentation and tracking methods to have the capacity to train fully supervised segmentation methods. 
    \item We conduct extensive experiments and demonstrate that our method is competitive with the state-of-the-art system~\cite{liu2021weakly} and outperforms fully supervised MaskTrack R-CNN~\cite{yang2019video} and TrackR-CNN~\cite{voigtlaender2019mots}.
\end{itemize}

\section{Related Work}
Although weakly supervised instance segmentation in videos is relatively under-studied, this  section summarizes the recent advances in the related fields regarding weakly supervised instance segmentation, box-supervised methods, and segmenting in videos \cite{cholakkal2019object,ge2019label,hsu2019weakly,tian2021boxinst,zhou2018weakly,zhu2019learning}.

\subsection{Fully Supervised Instance Segmentation} 
In the past decade, various fully supervised image instance segmentation methods have been proposed. These approaches can generally be divided into two categories: two-stage and single-stage approaches. Two-stage methods \cite{he2017mask,liu2018path,huang2019mask,chen2019hybrid} typically generate multiple object proposals in the first stage and predict masks in the second stage. While two-stage methods achieve high accuracy with large computational cost, single-stage approaches \cite{bolya2019yolact,pinheiro2016learning,dai2016instance,peng2020deep,Cao2020SipMaskSI,chen2020blendmask} employ predictions of bounding-boxes and instance masks at the same time. For example, SipMask~\cite{Cao2020SipMaskSI} proposes a novel light-weight spatial preservation module that preserves the spatial information within a bounding-box. BlendMask~\cite{chen2020blendmask} is based on the fully convolutional one-stage object detector (FCOS) \cite{tian2019fcos}, incorporating rich instance-level information with accurate dense pixel features. However, all these methods are built upon accurate human-labeled mask annotations, which requires far more human annotators than box annotations. In contrast, our method uses only box annotations instead of mask annotations, and thus dramatically reduces labeling efforts.

\subsection{Weakly-supervised Instance Segmentation} 
Using class labels to extract masks from CAMs or similar attention maps has gained popularity in training weakly supervised instance segmentation models \cite{cholakkal2019object,ge2019label,shen2019cyclic,zhou2018weakly}.  However, CAM-based supervision is not intrinsically suitable for the instance segmentation task as it cannot provide accurate information regarding individual objects, which potentially causes confusion in prediction~\cite{ahn2019weakly,chen2020banet,hur2016joint,wang2020self}. One closely related work is flowIRN~\cite{liu2021weakly}, which uses the flow fields as the extra supervision signal to operate training. Our technique is conceptually distinct in three folds: 1) flowIRN only uses flow field to generate pseudo-labels and thus fails to fully exploit the spatio-temporal representations. In contrast, we leverage the collaborative power of spatio-temporal collaboration to produce high-quality pseudo-labels; 2) flowIRN trains different contributing modules ($i.e.,$ CAM and optical flow) dis-jointly, resulting in an ineffective and complicated training pipeline. We propose a puzzle solver that organizes learning through the use of our pseudo-labels with box annotations, enable a fully end-to-end fashion; 3) flowIRN directly adopts exiting tracking method~\cite{voigtlaender2019mots}, while our tracking module builds on a novel diagonal-point-based approach. More comparison results will be provided in the experiments.
To address this issue, we aim to explore the more effective pseudo-labels from spatio-temporal collaboration for weak supervision.

\subsection{Box-supervised Instance Segmentation} 
Our work is also closely related to the box-supervised instance segmentation methods.
At the image level, SDI \cite{khoreva2017simple} might be the first box-supervised instance segmentation framework, which utilizes candidate proposals generated by MCG \cite{pont2016multiscale} to operate segmentation. In the same vein, a line of recent work \cite{arun2020weakly,hsu2019weakly,lee2021bbam,liu2020leveraging} formulates the box-supervised instance segmentation by sampling the positive and negative proposals based on the ROIs feature maps. However, using proposals for instance segmentation has redundant representations because a mask is repeatedly encoded at each foreground feature around ROIs. In contrast, our method is proposal-free as we remove the need for proposal sampling to supervise the mask learning. BoxInst \cite{tian2021boxinst} is one of the works that is similar to ours. It uses a pairwise loss function to operate training on low-level color features. However, their pairwise loss works in an oversimplified manner that encourages confident pixel neighbors to have similar mask predictions, inevitably introducing noisy supervision. Different from BoxInst, our method produces high-quality pseudo-labels derived from high-level spatio-temporal priors for supervision. To organize learning, we devise a novel puzzle loss to supervise our mask generation to capture accurate instance boundaries with box annotations.

\subsection{Video Segmentation} 
A series of fully supervised approaches have emerged for segmentation in videos \cite{athar2020stem,bolya2019yolact,liu2021sg,rajchl2016deepcut,xie2020polarmask,liu2020guided,liu2005combined,gui2019reliable}. For instance, VIS \cite{yang2019video} and MOTS \cite{voigtlaender2019mots} both extend Mask R-CNN \cite{he2017mask} from images to videos and simultaneously segment and track all object instances in videos. 
To the best of our knowledge, flowIRN \cite{liu2021weakly} and BTRA \cite{lin2021bilateral} may be two of the few that explore weakly supervised learning for the video-level instance segmentation task. FlowIRN \cite{liu2021weakly} trains different contributing modules ($i.e.,$ CAM and optical flow) dis-jointly and incurs additional dependencies, resulting in a dense training pipeline. BTRA \cite{lin2021bilateral} only box to generate pseudo-labels and thus fails to fully exploit the spatio-temporal representations for the boundary supervision. 
To maximize synergies for instance segmentation in videos, we propose a weakly supervised spatio-temporal collaboration framework in the paper. Unlike the aforementioned methods which overlook the sub-task of tracking in videos, we implement a strong tracking module to model instance movement across frames by using diagonal points with spatio-temporal information. Compared to the prior efforts \cite{liu2021weakly,voigtlaender2019mots,yang2019video}, our tracking module has a more robust tracking capacity.
\begin{figure*}[tb]
    \centering
    \includegraphics[width=\linewidth]{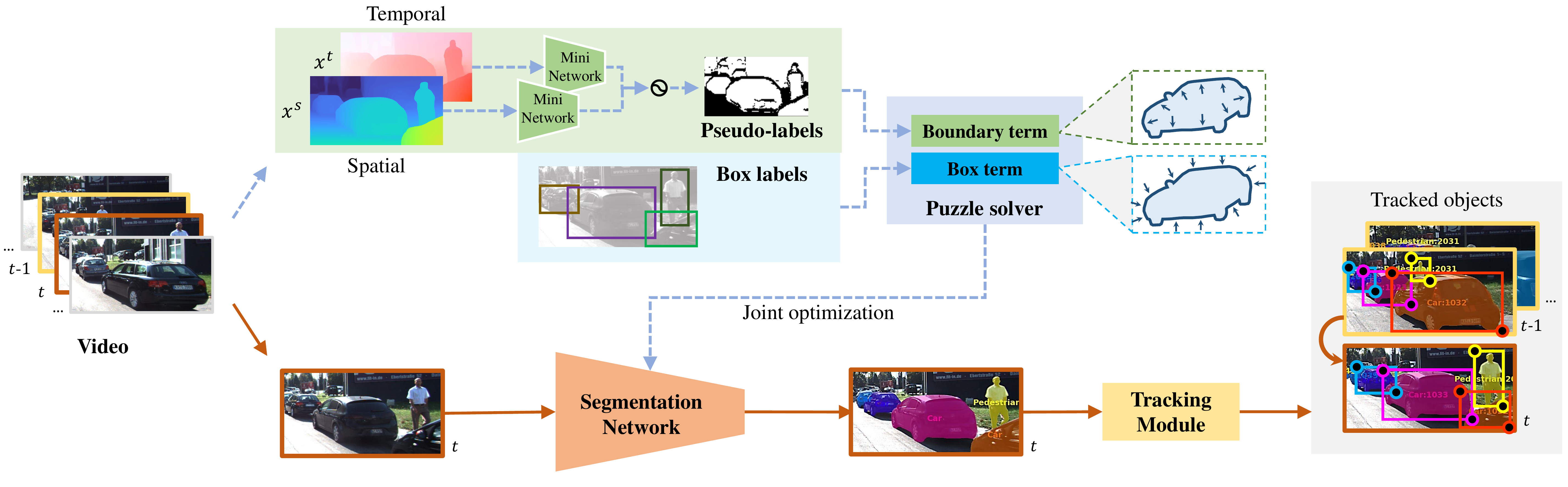}
    \caption{\textbf{The overview of STC-Seg framework}.
    During training, pseudo-labels from spatio-temporal collaboration and box labels from box annotation are jointly fed into the puzzle solver to learn a unified instance segmentation network.
    During inference, the learned segmentation network is applied to every frame, followed by a tracking module to perform robust object tracking. Dashed and solid paths are the pipelines for training and inference respectively.}
    \label{framework}
\end{figure*}

\subsection{Spatio-Temporal Collaboration}
A bunch of previous works \cite{yan2022video,shu2021spatiotemporal,shu2022expansion,yan2018GLRG,athar2020stem,cui2021tf,li2019spatio,qiu2017learning} explore spatio-temporal collaboration to assist visual tasks. For example, P3D ResNet \cite{qiu2017learning} mitigated limitations of deep 3D CNN by devising a family of bottleneck building blocks that leverages both spatial and temporal convolutional filters. 

SC-RNN \cite{shu2021spatiotemporal} simultaneously captures the spatial coherence and the temporal evolution in spatio-temporal space. ESE-FN \cite{shu2022expansion} captures motion trajectory and amplitude in spatio-temporal space using skeleton modality, which is effective in modeling elderly activities. However, these methods embed spatio-temporal analysis into the entire model, where the spatio-temporal modeling process is required during inference. In our method, the spatio-temporal collaboration is only used as the supervision signal during training, but not needed in the segmentation prediction.

\section{STC-Seg Approach}\label{STC-Seg}
\subsection{Overall Framework}
The overall framework of STC-Seg is shown in Fig.~\ref{framework}. During training, the pseudo-labels are first generated with spatio-temporal collaboration. The segmentation model is then jointly learned based on the pseudo-labels and the box labels/annotations via a novel puzzle solver.
During inference, we directly perform instance segmentation on input video data without using any extra information (\ie, depth estimation or optical flow).
Essentially, our STC-Seg consists of three core components: 1) the spatio-temporal pseudo-label generation, which offers a supervision signal for our training; 2) the puzzle solver, which organizes the training of video instance segmentation models; and 3) the tracking module, which enables robust tracking capacity. 
We present the details of each component in the following sections.

\subsection{Puzzle Solver with Spatio-temporal Collaboration}
\subsubsection{Pseudo-label Generation.}
\label{Pseudo-label-Generation}
Most existing works \cite{arun2020weakly,lee2021bbam,liu2020leveraging} rely solely on optical flow to generate pseudo-label. 
In this work, we leverage both spatial and temporal signals in our pseudo-label generation pipeline 
to better capture rich boundary information and effectively distinguish the foreground (the instance) from the background. In particular, our method adopts spatial signal $\bm{S}^s$ obtained from depth estimation~\cite{Godard2019DiggingIS}, and temporal signal $\bm{S}^t$ obtained from optical flow~\cite{luo2021upflow}.
 
As shown in Fig.~\ref{framework},
we directly employ depth estimation $\bm{x}^s$ $\in$ $\mathbb{R}^{h\times w\times 1}$ and optical flow $\bm{x}^t$ $\in$ $\mathbb{R}^{h\times w\times 2}$
as the inputs for our pseudo-label generation module. 
The above two inputs keep the same resolution $w\times h$ with the input frame, in order to build the pixel-to-pixel correspondence. Each signal $\bm{x} \in \{\bm{x}^s, \bm{x}^t\}$ is then fed into a mini network \cite{wei2018revisiting} to compute the contextual similarity at each pixel location for obtaining the spatial and temporal signals, respectively. Given a location $(i,j)$ on the input $\bm{x}$, the contextual similarity score $\bm{S}_{i,j}$ on the corresponding signal $\bm{S} \in \{\bm{S}^s, \bm{S}^t\}$ 
is computed as:
\begin{equation}
\begin{aligned}
    &\bm{S}_{i,j}=\sum_{k_1,k_2} \delta \left(\bm{w}_{k_1,k_2}\cdot \bm{x}_{\left(i+\lambda \cdot k_{1}\right),\left(j+\lambda \cdot k_{2}\right)}, \bm{x}_{i,j}\right)
    \label{eq:signal}
\end{aligned}
\end{equation}
where $k_1,k_2 \in \{-1,0,1\}$. $\bm{w}$ is the dilated kernel, and $\lambda$ is the dilation rate. $\bm{\delta(,)}$\footnote{$\delta\left(\bm{x}_{i,j}, \bm{x}_{i',j'}\right)=e^{r\cdot \left\vert\left\vert \bm{x}_{i,j}-\bm{x}_{i',j'} \right\vert\right\vert_p}$.} is the similarity measurement function. For the obtained signals, we have $\bm{S}^s, \bm{S}^t$ $\in$ $\mathbb{R}^{h\times w\times 1}$.

To produce the pseudo-label $\bm{M}$ for training, we leverage the complementary representations of the two signals by fusing them together with a threshold filter:
\begin{equation}
\begin{aligned}
    \bm{M}=(\bm{S}^s - \phi^s) \wedge (\bm{S}^t - \phi^t)
\end{aligned}
\end{equation}
where 
$\phi^s, \phi^t$ denote the filter factor to determine  the   salience threshold of each signal on the foreground instances.
However, noises may reside on the pseudo-labels and segregate one target instance into multiple sub-regions.

\begin{figure*}[tb]
    \centering
    \includegraphics[width=\linewidth]{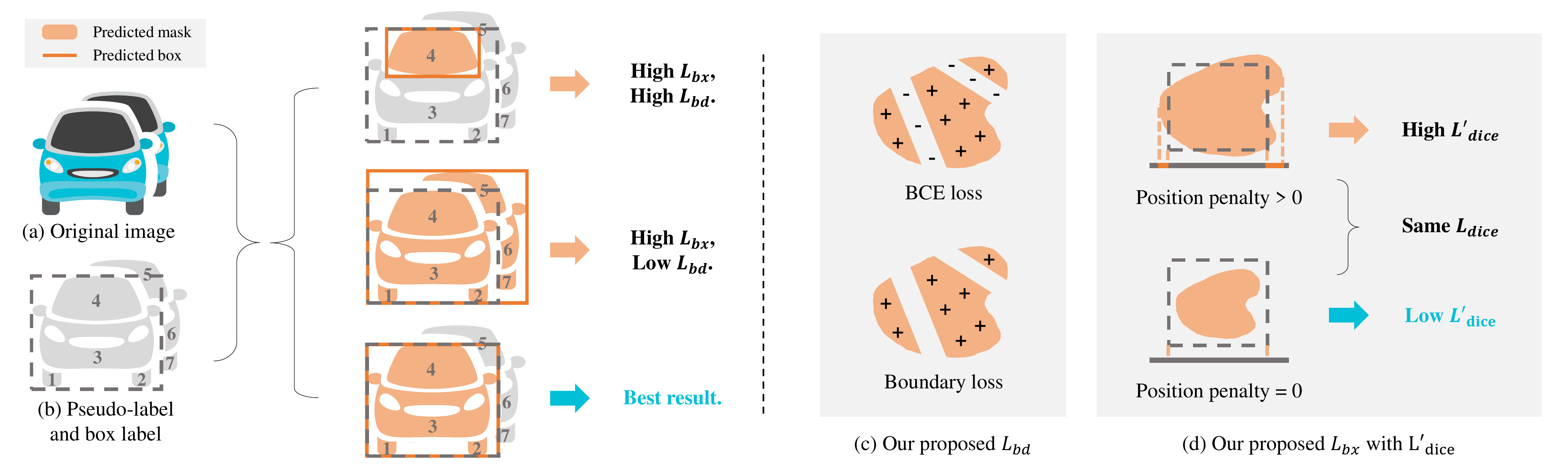}
    \caption{\textbf{Demonstration of puzzle solver}. Our puzzle solver performs strong supervision with box annotations and pseudo labels. Albeit the corresponding pseudo labels for one target generally include multiple sub-regions, \ie, sub-region 1-4 in (b), 
    the boundary term and box term in our puzzle loss work collaboratively 
    to supervise the mask prediction for aligning the shape of the instance while being consistent with the ground truth box. 
    }
    \label{fig:loss_explanation}
    \vspace{-8pt}
\end{figure*}

\subsubsection{Puzzle Solver.}



As mentioned above, directly using the pseudo-labels without constraints may result in excessively noisy supervision and suboptimal training outcomes. In comparison to fully supervised information, which can be labeled pixel-by-pixel, solving the puzzle of predicting the imaginary mask is difficult in the weakly supervised fashion. To address this issue, we introduce a novel puzzle solver that organizes learning through the use of our pseudo-labels with box annotations. 

Our puzzle solver essentially designs a puzzle loss that operates supervision of mask prediction with two loss terms. The first one is \textit{Boundary term}, which explores all the candidate sub-regions of the target instances to depict their boundaries. The second one is \textit{Box term}, which ensures maximal positions of the predicted mask boundaries can closely stay within the ground truths. \textit{The two terms work collaboratively to solve the puzzle of how to assemble  suitable sub-region masks together to  match the shape of the instance} (see Fig.~\ref{fig:loss_explanation}). Our puzzle solver is to jointly optimize both the boundary term $L_{bd}$ and box term $L_{bx}$ with respect to the network parameters $\theta$:
\begin{equation}
    \begin{aligned}
        \mathop{\arg\min}_{\theta} L_{pz} = \mathop{\arg\min}_{\theta} ( L_{bd}+L_{bx})
    \end{aligned}
\end{equation}

\textit{Boundary term:}
With ground truths, fully supervised methods can use binary cross entropy (BCE) loss $L_{bce}$ to supervise the mask generation, which uses both positive samples (the foreground)  and negative samples (the background) in training. However, as discussed in Section
\ref{Pseudo-label-Generation}, our pseudo-labels are noisy references with unwanted inner negative samples inside the object, 
which would introduce inevitable noises in training.  To address this issue, we modify $L_{bce}$ by focusing on learning positive examples to capture the instance boundaries (Fig.~\ref{fig:loss_explanation}c). Concretely, given a pixel location $(i, j)$ on the pseudo-labels $\bm{M}$, its corresponding label $m_{i,j}$ can be $m_{i,j} \in \{0,1\}$, where 1 denotes the foreground instance and 0 denotes the background. To learn the instance mask generation, our boundary loss only operates learning of the posterior probability $P(\Tilde{m}_{i,j} | m_{i,j}=1)$ from positive samples, where $\Tilde{m}_{i,j} \in \{0, 1\}$ is the predicted mask at $(i, j)$. Given the input size $w\times h$, our boundary loss is given by:
\begin{equation}
    \begin{aligned}
        L_{bd} = -\frac{1}{h \times w} \sum_{j=1}^{w}\sum_{i=1}^{h} m_{i,j} \log P(\Tilde{m}_{i,j}=1)
    \end{aligned}
\end{equation}
\indent At first glance, only using the positive sampling may not work well in training. However, an important observation is that our pseudo-labels allow the network to effectively learn the dominant representations from the positive examples. With additional box supervision, the boundary loss computation effectively captures the instance boundaries, thus largely eliminating supervision noises.

\textit{Box term:} 
To perform box-level supervision, BoxInst \cite{tian2021boxinst} adopts dice loss  \cite{Milletari2016VNetFC}, which computes the similarity distance of the predicted bounding boxes and the ground truths. However, 
a prediction, which is larger or smaller than the ground truth, may have a similar penalty in dice loss computation. Thus, a model supervised by the dice loss tends to generate overly saturated masks that go beyond the boxes.
To address this issue, we introduce a position penalty into dice loss $L_{dice}$ to penalize the model for generating a mask that exceeds the box (as shown in Fig.~\ref{fig:loss_explanation}d). This penalty term encourages the mask boundary to align with the ground truth box:
\begin{equation}
    \begin{aligned}
        L^{'}_{dice}(p,g)
        &= \underbrace{\frac{2\sum\limits_i^N p_{i} g_{i}}{\sum\limits_i^N p_{i}^2 
        +g_{i}^2}}_{\text{Dice loss}~L_{dice}} + \underbrace{\frac{\sum\limits_i^N [\max(p_{i}-g_{i}, 0)]^2}{\sum\limits_i^N g_{i}^2}}_{\text{Position penalty}}
    \end{aligned}
    \label{eq:penalty}
\end{equation}
where $p_{i} \in (0,1)$ and $g_{i} \in \{0,1\}$ are the log-likelihood scores of the prediction and the ground truth respectively. $N$ is the length of the input sequence.
The position penalty can be understood as the proportion of the predicted region that exceeds the ground truth region. As shown in Eq.~\ref{eq:penalty}, it is clear that there is no position penalty for those points within the ground truth.
The final box term can be written as:
\begin{equation}
    \begin{aligned}
        L_{bx}(\Tilde{m}, B) = L_{dice}^{'}(\mathrm{Proj}_x(\Tilde{m}), \mathrm{Proj}_x(B)) \\
        + L_{dice}^{'}(\mathrm{Proj}_y(\Tilde{m}), \mathrm{Proj}_y(B)) 
    \end{aligned}
\end{equation}
where $\Tilde{m}$ is the predicted instance mask. $B$  is the corresponding box annotations. $\mathrm{Proj}_x$ and $\mathrm{Proj}_y$ are the projection functions \cite{tian2021boxinst}, which map $\Tilde{m}$ and $B$ onto $x$-axis and $y$-axis, respectively. It is worth mentioning that the new 
$L_{bx}$ effectively rectifies the expanded masks outside the box that are introduced by the $L_{bd}$. In other words,
the $L_{bd}$ allows the model to predict larger masks, while $L_{bx}$ ensures the model predicts precise masks that are consistent with the ground truth boxes.
Note that the segmentation generation is independent of pseudo-label generation. The computational cost only increases when calculating the losses, which has the same computational complexity as the MSE loss. Therefore, our method will not introduce additional computation cost.

Our method can also be applied to those tasks with noise supervision, such as target segmentation tasks with inaccurate box labeling or incorrect labels \cite{xu2021training,shu2019meta,lee2018cleannet,li2020learning}. For the former cases, we can slightly modify the loss of box term to assign a larger weight to the positive feedback of the intersection area, while assigning the negative feedback outside the intersection area a smaller weight. For the latter cases, we can modify the loss of boundary term by assigning a relatively large weight to the positive item and a small weight to the negative item in the cross entropy. In this way, our loss function is able to deal with more inaccurate box annotations and label predictions. Intuitively, the classification task can be regarded as a regression problem by taking the irrelevant labels as the "background", so that the regression boundary can shrink inward on the feature plane until the accurate label boundary is found.



\subsection{Tracking Module}\label{tra}
Existing methods~\cite{luiten2021hota,yin2021center,Zhou2020TrackingOA} prioritize object position modeling for tracking, which may cause confusion when two objects are extremely occluded or overlapped as shown in Fig.~\ref{framework}. To address this issue, we place a premium on both object size and position modeling in our tracking module. Moreover, the spatio-temporal changes on individual objects should remain within a reasonable range, given the consistency of video object movement across frames. In light of both observations, we introduce a novel tracking module using \textit{diagonal points} with \textit{spatio-temporal discrepancy}.

\subsubsection{Diagonal Points Tracking} 
To represent the object position and size, we adopt diagonal points to model the object movement by using
the upper-left corner $(x_1, y_1)$ and the lower-right corner $(x_2, y_2)$ of the bounding box. Similar to almost tracking methodology\cite{yin2021center,Wojke2017SimpleOA}, we adopt a recursive \textit{Kalman Filter} and frame-by-frame data association to predict the future location for each tracked object. The movement $\Delta l_o^{t-1\rightarrow t}$ of a tracked object $o$ in the $t_{th}$ frame is used to predict the future location $p^{t+1}(l_o^{t})=\Delta l_o^{t-1\rightarrow t} + l_o^{t}$ of this object at $(t+1)_{th}$, where $l_o^{t}$ is the location of object $o$ and $\Delta l_o^{t-1\rightarrow t}$ is given by:
\begin{equation}
\begin{aligned}
    &\Delta l_o^{t-1\rightarrow t}=(x_1^t-x_1^{t-1}, y_1^t-y_1^{t-1}, x_2^{t}-x_2^{t-1}, y_2^{t}-y_2^{t-1}) 
\end{aligned}
\end{equation}
\indent During object tracking, we maintain a dictionary $O_{\leq t}=\{\hat{o}\}^{\hat{K}}$ of $\hat{K}$ tracked objects in former frames. Given $K$ detected objects $O_{t+1}=\{o\}^K$ in $(t+1)_{th}$ frame, our tracking is to build a list of one-to-one matching pairs $\hat{o} = \varphi (o) \in O_{\leq t}$ to minimize the Euclidean distance between ground truth locations $l_o^{t+1}$ of each $o \in O_{t+1}$ and the predicted future locations $p^{t+1}(l_{\varphi(o)}^{t})$:
\begin{equation}
\begin{aligned}
    \mathop{\arg\min}\limits_{\varphi} \sum_{O_{t+1}}<p^{t+1}(l_{\varphi(o)}^{t}),l_o^{t+1}>
\end{aligned}
\label{eq:tracking}
\end{equation}

\subsubsection{Bi-greedy Matching}
Conventional tracking methods are generally one-directional as they perform a popular greedy search, called \textit{Hungarian Algorithm},  to build the correspondences $\varphi$ from the previous  frame to the current one (\eg, JDE\cite{wang2020towards}, DeepSORT\cite{Wojke2017SimpleOA}, FairMot\cite{zhang2021fairmot}, CenterTrack\cite{Zhou2020TrackingOA}). However, the position of the same object in the previous frame may not always be the closest one that appeared in the current frame, which causes confusion in tracking. To address this problem, some methods use pre-trained CNN descriptors to distinguish objects \cite{yang2019video,Wojke2017SimpleOA}, but computing features takes too much time and the objects are sometimes very similar. Thus, we consider matching from both directions (i.e., previous-to-current and current-to-previous) and develop a bidirectional greedy matching to output the tracking $T_{\varphi}$ as follows (assuming that only DP is used):

\begin{algorithm}[H]
  \renewcommand{\algorithmicrequire}{\textbf{Input:}}
  \renewcommand{\algorithmicensure}{\textbf{Output:}}
  
  \caption{Bi-greedy matching.}
  \label{Bi-Greedy_Matching}
  
  \begin{algorithmic}[1]
      \REQUIRE $O_{\leq t}=\{\hat{o}\}^G$, $O_{t+1}=\{o\}^K$. 
      \ENSURE $T_{\varphi} = \{(o, \varphi(o))\}^K$ for all $o \in O_{t+1}$
      \STATE $T \leftarrow \emptyset$
      \STATE $T_{\varphi} \leftarrow \emptyset$\
      \FORALL{$\hat{o} \in O_{\leq t}$}
      \STATE $ o' \leftarrow \mathop{\arg\min}\limits_{o \in O_{t+1}} <p^{t+1}(l_{\hat{o}}^{t}),l_o^{t+1}> $
      \STATE $ T \leftarrow T \cup (\hat{o}, o')$ 
      \ENDFOR
      \FORALL{$o \in O_{t+1}$}
      \IF{any $ \hat{o}, (\hat{o},o) \in T$}
      \STATE $\hat{o} \leftarrow \mathop{\arg\min}\limits_{\hat{o},(\hat{o},o) \in T} <p^{t+1}(l_{\hat{o}}^{t}),l_o^{t+1}> $ 
      \STATE $ T_{\varphi} \leftarrow T_{\varphi} \cup (o, \hat{o})$ 
     \ELSE 
     \STATE $T_{\varphi} \leftarrow T_{\varphi} \cup (o, New \, \hat{o})$
     \ENDIF
     \ENDFOR
      
  \end{algorithmic}
\end{algorithm}

As shown in Algorithm~\ref{Bi-Greedy_Matching}, our proposed matching algorithm first finds the nearest instance $o' \in O_{t+1}$ in the current frame for each previous tracked object $\hat{o} \in O_{\leq t}$. There may exist two cases: (a) more than one different previous instance may have the same nearest current instance $o$, those previous instances are collected as \textit{candidate instances}; (b) it is also possible that some current instances are not marked by any previous instances. In the case (a), for this current instance $o$, we finally get the matched previous instance $\hat{o}$ by finding the nearest one from those \textit{candidate instances}. In the case (b), those current instances are judged as a new instance. Since $G>K$ in most cases, to run traversal fist on $O_{\leq t}$ is better than $O_{t+1}$ because it focuses on the matched current instances $o' \in O_{t+1}$ rather than previous instances that are not in current frame. In each round of matching, in order to avoid the occluded objects being forgotten, we need to re-match the newly emerged objects. Therefore, we adopt a matching cascade algorithm \cite{Wojke2017SimpleOA} that gives priority to more frequently appearing objects to ensure those objects that are briefly occluded and disappeared can be re-identified.

\subsubsection{Occluded Object Multi-stage Matching} 
The occluded objects often get a low confidence level after passing through the detection algorithm. The existing algorithms only set a single confidence level threshold to divide the correctly detected target and the wrongly detected target. This approach causes the tracking of occluded objects to fail. We introduce a \textit{multi-stage matching} mechanism, that is to set two lower thresholds of confidence scores, and treat the divided high-scoring detection targets and low-scoring detection targets differently, and perform two rounds of matching respectively in turn. In this way, although the confidence obtained by the occluded object position is lower, it can still be successfully matched in the second round of matching.

\subsubsection{Spatio-Temporal Discrepancy}  Considering the fact that the spatio-temporal changes on individual objects should retain a reasonable range in videos, we extend the Eq.~\ref{eq:tracking} by adding the spatio-temporal discrepancy for tracking:
\begin{equation}
\begin{aligned}
    \mathop{\arg\min}\limits_{\tau} \sum_{O_{t+1}} 
    \alpha_1 <p^{t+1}(l_{\varphi(o)}^{t}),l_o^{t+1}> \\
    + \alpha_2<D^{t}(l_{\varphi(o)}^{t}),D^{(t+1)}(l_o^{t+1})> \\
    + \alpha_3<F^{t}(l_{\varphi(o)}^{t}),F^{(t+1)}(l_o^{t+1})>
\end{aligned}
\label{eq:tracking1}
\end{equation}
where $D^{t}$ and $F^{t}$ denotes the depth and optical flow values 
of the diagonal points for the tracked object $o$
on the $t_{th}$ frame. $\alpha_1, \alpha_2, \alpha_3$ are the trade-off weights that balance these terms. The new objective essentially ensures the tracked objects are aligned with the segmented instances among frames, while at the same time being consistent with their spatio-temporal positions.
We demonstrate the improvements of our tracking in Section~\ref{Understanding_STC_Seg}.
\vspace{-5pt}
\section{Experiments}\label{Experiments}
\vspace{-1pt}
\begin{table*}[t]
\caption{\textbf{Quantitative results on KITTI MOTS test set.} Results for fully supervised methods are retrieved from the MOTS benchmark. For weakly supervised methods WISE and IRN, the results are obtained from their original codes combined with our tracking module. STC-Seg$_{50}$ and STC-Seg$_{101}$ indicate using ResNet-50 and ResNet-101 as backbone respectively. All the baseline methods use ResNet-101 with FPN.}
    \label{tab:mots_results}
    \centering
    
    \begin{tabular}{c|c|cccc|cccc}
    
    \hlineB{2.5}
	\rowcolor{mygray} 
    &  & \multicolumn{4}{c|}{Car} & \multicolumn{4}{c}{Pedestrian} \\
    \rowcolor{mygray} 
    & \multirow{-2}{*}{Methods}  & HOTA & sMOTSA & MOTSA & MOTSP & HOTA & sMOTSA & MOTSA & MOTSP  \\
    
    \hline
    \midrule
    
    \multirow{5}{*}{\rotatebox{90}{Fully}}
    & ViP-DeepLab \cite{vip_deeplab}         & 76.3 & 81.0 & 90.7 & 89.8 & 64.3 & 68.7 & 84.5 & 82.3 \\
    & EagerMOT \cite{Kim21ICRA}              & 74.6 & 74.5 & 83.5 & 89.5 & 57.6 & 58.0 & 72.0 & 81.5 \\
    & MOTSFusion \cite{luiten2019MOTSFusion} & 73.6 & 74.9 & 84.1 & 89.3 & 54.0 & 58.7 & 72.8 & 81.5 \\
    & PointTrack \cite{xu2020Segment}        & 61.9 & 78.5 & 90.8 & 87.1 & 54.4 & 61.4 & 76.5 & 80.9 \\
    & TrackR-CNN \cite{voigtlaender2019mots} & 56.6 & 66.9 & 79.6 & 85.0 & 41.9 & 47.3 & 66.1 & 74.6 \\
    
    \midrule
    \multirow{4}{*}{\rotatebox{90}{Weakly}}
    & WISE \cite{laradji2019masks} & 41.8 & 21.6 & 39.5 & 62.9 & 20.9 & 18.8 & 29.1 & 55.2 \\
    & IRN \cite{ahn2019weakly} & 44.7 & 25.9 & 41.1 & 64.1 & 22.9 & 19.1 & 31.6 & 56.4 \\ 
    & FlowIRN \cite{liu2021weakly} & 50.2 & 45.1 & 63.8 & 71.4 & 27.5 & 19.4 & 35.9 & 62.7 \\
    & PointRend \cite{cheng2022pointly} & 51.8 & 49.3 & 70.6 & 74.2 & 28.3 & 20.0 & 37.5 & 64.4 \\
    & MOTSNet + Grad-CAM  \cite{ruiz2021weakly} & - & 54.6 & 72.5 & 76.6 & - & 20.3 & 39.7 & 65.7 \\
    & \textbf{STC-Seg$_{50}$} & 57.7 & 66.9 & 80.7 & 83.9 & 46.7 & 47.4 & 67.6 & 73.6  \\
    & \textbf{STC-Seg$_{101}$} & \textbf{59.6} & \textbf{69.2} & \textbf{83.3} & \textbf{85.1} & \textbf{47.5} & \textbf{48.6} & \textbf{68.3} & \textbf{75.8} \\
    
    
    
    
    \bottomrule

    \end{tabular}
    
\end{table*}
\begin{table*}[tbp]
\caption{\textbf{Results on YT-VIS validation set.} Metrics for SipMask \cite{Cao2020SipMaskSI} are obtained from its original paper. All other compared results are retrieved from \cite{liu2021weakly}. All methods use ResNet-50 with FPN.}
    \label{tab:yt_vis_results}
    \centering
    
    \begin{tabular}{c|c|c|c|c|c|c}
         \hlineB{2.5}
         \rowcolor{mygray} 
         \multicolumn{2}{c|}{Methods} & mAP & AP@0.5 & AP@0.75 & AR@1 & AR@10 \\
         
         \hline
         \midrule
         
         \multirow{4}{*}{\rotatebox{90}{Fully}}
          & IoUTracker+  \cite{yang2019video} & 23.6 & 39.2 & 25.5 & 26.2 & 30.9 \\
          & DeepSORT \cite{Wojke2017SimpleOA} & 26.1 & 42.9 & 26.1 & 27.8 & 31.3  \\
          & MaskTrack R-CNN \cite{yang2019video} & 30.3 & 51.1 & 32.6 & 31.0 & 35.5  \\
          & SipMask \cite{Cao2020SipMaskSI} & 33.7 & 54.1 & 35.8 & 35.4 & 40.1 \\
         \midrule
         \multirow{4}{*}{\rotatebox{90}{Weakly}}
          & WISE \cite{laradji2019masks} &  6.3 & 17.5 & 3.5 & 7.1 & 7.8  \\
          & IRN \cite{ahn2019weakly}  & 7.3 & 18.0 & 3.0 & 9.0 & 10.7  \\
          & FlowIRN \cite{liu2021weakly} & 10.5 & 27.2 & 6.2 & 12.3 & 13.6  \\
          & \textbf{STC-Seg$_{50}$} & \textbf{31.0} & \textbf{52.4} & \textbf{33.2} & \textbf{32.9} & \textbf{36.2} \\
          \bottomrule
    \end{tabular}
    
\end{table*}

\subsection{Datasets} We evaluate STC-Seg on two benchmarks: KITTI MOTS \cite{voigtlaender2019mots} and YT-VIS \cite{yang2019video}. The KITTI MOTS contains 21 videos (12 for training and 9 for validation) focusing on driving scenes. The YT-VIS contains 2,883 YouTube video clips with 131k object instances and 40 categories. On KITTI MOTS, the metrics are HOTA, sMOTSA, MOTSA, and MOTSP from \cite{luiten2021hota}. On YT-VIS, the metrics are: mAP is the mean average precision for IoU between [0.5, 0.9], AP@0.50 and AP@0.75 are average precision with IoU threshold at 0.50 and 0.75, and AR@1 and AR@10 are average recall for top 1 and 10 respectively.

\begin{figure}[t]
    \centering
    \includegraphics[width=1.0\linewidth]{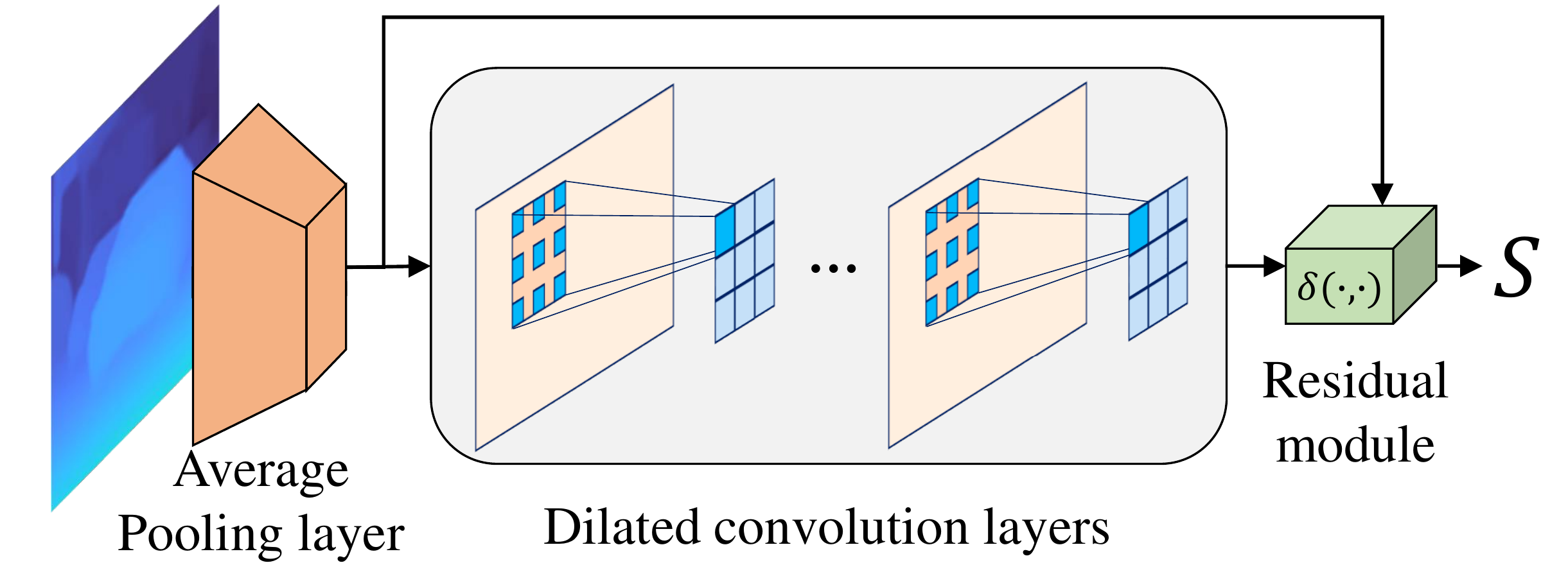}
    \caption{The architecture of our mini network. We use the dilated convolution layers to capture the spatial or temporal data difference between adjacent pixels. The similarity measurement function $\delta(\cdot,\cdot)$ is implemented by a residual module.}
    \label{fig:mini_network}
\end{figure}

\subsection{Pseudo-Label Generation Network Architecture} In the pseudo-label generation of STC-Seg, unsupervised Monodepth2 \cite{Godard2019DiggingIS} and Upflow \cite{luo2021upflow} are adopted for the depth estimation and optical flow respectively. The depth and flow outputs are fed into a mini network. The mini network is composed of a 2D average pooling layer, dilated convolution layers and a residual module, as shown in Fig.~\ref{fig:mini_network}. In pre-processing, we use a 2D average pooling layer to down-sample the depth and the optical flow data. The kernel size and the stride of the pooling layer are both set to 4 without padding. After the 2D average pooling layer, dilated convolution is applied since it enables networks to have larger receptive fields with just a few layers. The dilation rate $\lambda$ is set to 2 and the kernel size is set to 3 in our experiments, so the padding size is set to 2 to keep the output size equal to the input size. The weight of the kernel is initialized as $W = [w_{k_1,k_2}]_{3\times3}$, where $w_{k_1,k_2} \in \{0,1\}$. 
After the dilated convolution layers, the residual module subtracts the output of the pooling layer from the output of the dilated convolution layers and applies an exponential activation function. The final output of the residual module can be written as $\delta(\bm{x}_{i,j}, \bm{x}_{i',j'})=e^{r\cdot || \bm{x}_{i,j}-\bm{x}_{i',j'} ||_p}$ in Eq.~1, which represents the contextual similarity between locations $(i,j)$ and $(i',j')$ in the frame, where $r$ is the similarity factor. We use the Frobenius norm ($p=2$) in this contextual similarity calculation. The similarity factor $r$ is set to 0.5 in our experiments. 

\subsection{Main Network Architecture} 
Our segmentation network is crafted on CondInst \cite{Tian2020ConditionalCF} with a few modifications. Following CondInst, we use the FCOS-based network, which includes ResNet-50/101 backbones~\cite{he2016deep} with FPN~\cite{lin2017feature}, a detection built on FCOS, and dynamic mask heads. 
For the dynamic mask heads, we use three convolution layers as in CondInst,  but we increase the channels from 8 to 16 as in~\cite{tian2021boxinst}, which results in better performance with an affordable computational overhead. Without any network parameter consumption, our tracking module directly performs tracking over the output of the segmentation network.

\subsection{Implementation Details} 
\subsubsection{Pseudo-Label Generation}
To generate pseudo-labels, 
unsupervised Monodepth2 \cite{Godard2019DiggingIS} and Upflow~\cite{luo2021upflow} are adopted for the depth estimation and optical flow respectively. 
The Monodepth2 \cite{Godard2019DiggingIS} is trained on the KITTI stereo dataset \cite{geiger2012we} when we take experiments on KITTI MOTS. When using the monocular sequences in KITTI stereo dataset for training, we follow Zhou et al.’s \cite{zhou2017unsupervised} pre-processing to remove static
frames. This results in 39,810 monocular triplets (three temporally adjacent frames) for training and 4,424 for validation.
We use a learning rate of $10^{-4}$ for the first 15 epochs which is then dropped to $10^{-5}$ for the remainder. When we take experiments on YT-VIS, the model is pre-trained on NYU Depth dataset \cite{silberman2012indoor} with a learning rate $10^{-4}$. Following \cite{ranftl2020towards}, images are flipped horizontally with a 50\% chance, and randomly cropped and resized to 384 × 384 to augment the data and maintain the aspect ratio across different input images. Monodepth2 is finetuned on the YT-VIS with a learning rate of $10^{-5}$ and an exponential decay rate of $\beta_1=0.9, \beta_2=0.999$.
The Upflow \cite{luo2021upflow} is trained on KITTI scene flow dataset \cite{menze2015object} when we take experiments on KITTI MOTS. KITTI scene flow dataset \cite{menze2015object} consists of 28,058 image pairs ($t_{th}$ frame and $(t-1)_{th}$ frame). Following \cite{liu2020learning}, the learning rate is set to $10^{-4}$ and the Adam optimizer is used during training. When we take experiments on YT-VIS, the Upflow \cite{luo2021upflow} is pre-trained on FlyingThings \cite{MIFDB16} for 100k iterations with a batch size of 12, then trained for 100k iterations on FlyingThings3D \cite{MIFDB16} with a batch size of 6. The learning rate of the above two stages is both set to $1.2 \times 10^{-4}$. The model is finetuned on YT-VIS for another 100k iteration with a batch size of 6
and a learning rate of $10^{-4}$.
Our mini network includes 3 layers of dilated convolutions. For Eq.\ref{eq:signal}, the dilation rate $\lambda$ is set to be 2 and the kernel size of dilation convolution is set to be 3. The filter factors $\phi^s, \phi^t$ are set to be 0.3 and 0.4 respectively.

\subsubsection{STC-Seg Training and Inference}
The STC-Seg is implemented using PyTorch. It is trained with batch size 8 using 4 NVIDIA GeForce GTX 2080 Ti GPUs (2 images per GPU) with 16 workers.
During training, the backbone is pre-trained on ImageNet \cite{Deng2009ImageNetAL}. The newly added layers are initialized as in FCOS \cite{tian2019fcos}.
Following CondInst, the input images are resized to have a shorter side [640,~800] and a longer side at a maximum of 1333. The same data augmentation in CondInst \cite{Tian2020ConditionalCF} is used as well.
For KITTI MOTS, we remove 485 frames without any usable annotation so there are 4510 frames left for training. For YTVIS, there are 61341 frames used for training in total. Only left-right flipping is used as the data augmentation during training. Following CondInst \cite{Tian2020ConditionalCF}, the output mask is up-sampled to 1/4 resolution of the input image, and we only compute the loss for top 64 mask proposals per image.  
For optimization, we use a multi-step learning rate scheduler with a warm-up strategy in the first epoch. In our multi-step learning rate schedule, the base learning rate is set to be $10^{-4}$, which starts to decay exponentially after a certain number of iterations up to $2\times10^{-5}$. In the warm-up epoch, the learning rate is increased linearly from 0 to the base learning rate. The base learning rate is set to be $10^{-4}$, which starts to decay exponentially after a certain number of iterations up to $2\times10^{-5}$. The exact number of iterations varies for each setting as follows: (a) KITTI-MOTS: 10k total iterations, decay begins after 5k iterations; (b) YouTube-VIS: 80k total iterations, decay begins after 30k iterations. The momentum is set to 0.9. The weight decay is set to $10^{-4}$, while it is not applied to parameters of normalization layers. 
In inference, we can directly perform instance segmentation on input video data \textit{without using any extra information}. The hyper-parameters $\alpha_1, \alpha_2, \alpha_3$ are set to be 0.7, 0.2, and 0.1 respectively.

\subsection{Main Results}
\begin{figure*}[t]
    \centering
    \includegraphics[width=\linewidth, trim={0 0.5cm 0 0.5cm},clip]{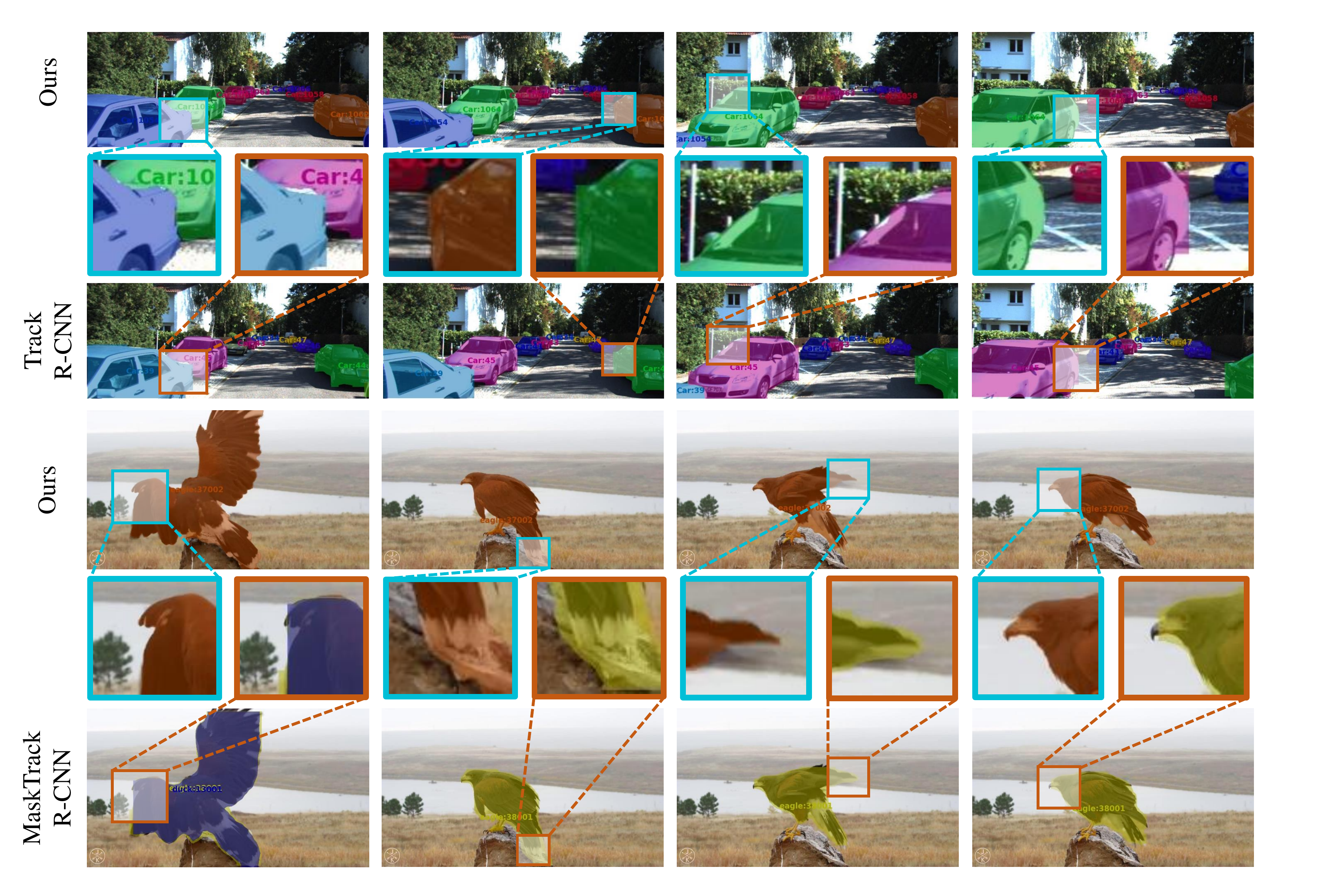}
    \caption{\textbf{Qualitative results} of our STC-Seg in comparison with TrackR-CNN \cite{voigtlaender2019mots} and MaskTrack
          R-CNN \cite{yang2019video} on KITTI MOTS and YT-VIS respectively. All compared methods use ResNet-101 with FPN.}
    \label{fig:Qualitative_result}
    \vspace{-8pt}
\end{figure*}
\subsubsection{Quantitative Results}
On the KITTI MOTS benchmark, we compare our STC-Seg against the state-of-the-art baselines. The results are presented in Table \ref{tab:mots_results}. It can be seen that our methods achieve competitive results under all evaluation metrics. Our STC-Seg with ResNet-50 significantly outperforms all weakly supervised methods which use a stronger backbone (ResNet-101). In comparison with the fully supervised methods, our method with
ResNet-101 can still achieve reasonable results. For example, it outperforms TrackR-CNN \cite{voigtlaender2019mots} by 3.0\% on HOTA, 2.3\% on sMOTSA, 3.7\% on MOTSA and 0.1\% on MOTSP for the car class. The results for pedestrian class also are consistent.
We further provide comparison results of our STC-Seg with the state-of-the-art baselines on YT-VIS in Table \ref{tab:yt_vis_results}. It can be seen that our method is competitive with fully supervised MaskTrack R-CNN~\cite{yang2019video} and SipMask \cite{Cao2020SipMaskSI}.
When comparing with weakly supervised methods, our method outperforms FlowIRN \cite{liu2021weakly}, IRN \cite{ahn2019weakly} and WISE \cite{laradji2019masks} with significant margins of 20.5\%, 23.7\%, and 24.7\% in terms of mAP metrics respectively.

\subsubsection{Qualitative Results} 
We compare qualitative results of our method with those from fully supervised TrackR-CNN \cite{voigtlaender2019mots} and MaskTrack R-CNN~\cite{yang2019video} on KITTI MOTS and YT-VIS respectively. To demonstrate the advantages of our approach, we select some challenging samples where TrackR-CNN and MaskTrack R-CNN have weaker predictions (see Fig. \ref{fig:Qualitative_result}). In the KITTI MOTS examples, the masks generated by Track RCNN have jagged boundaries or leave
false negative regions on the borders. In the YT-VIS examples, MaskTrack
R-CNN struggles to depict the boundary of instances with irregular shapes (\eg, eagle beak or tail). 
On the other hand, it is clear that our method captures more accurate instance boundaries.

\subsubsection{Discussion}
\begin{figure}[t]
    \centering
    \includegraphics[width=\linewidth]{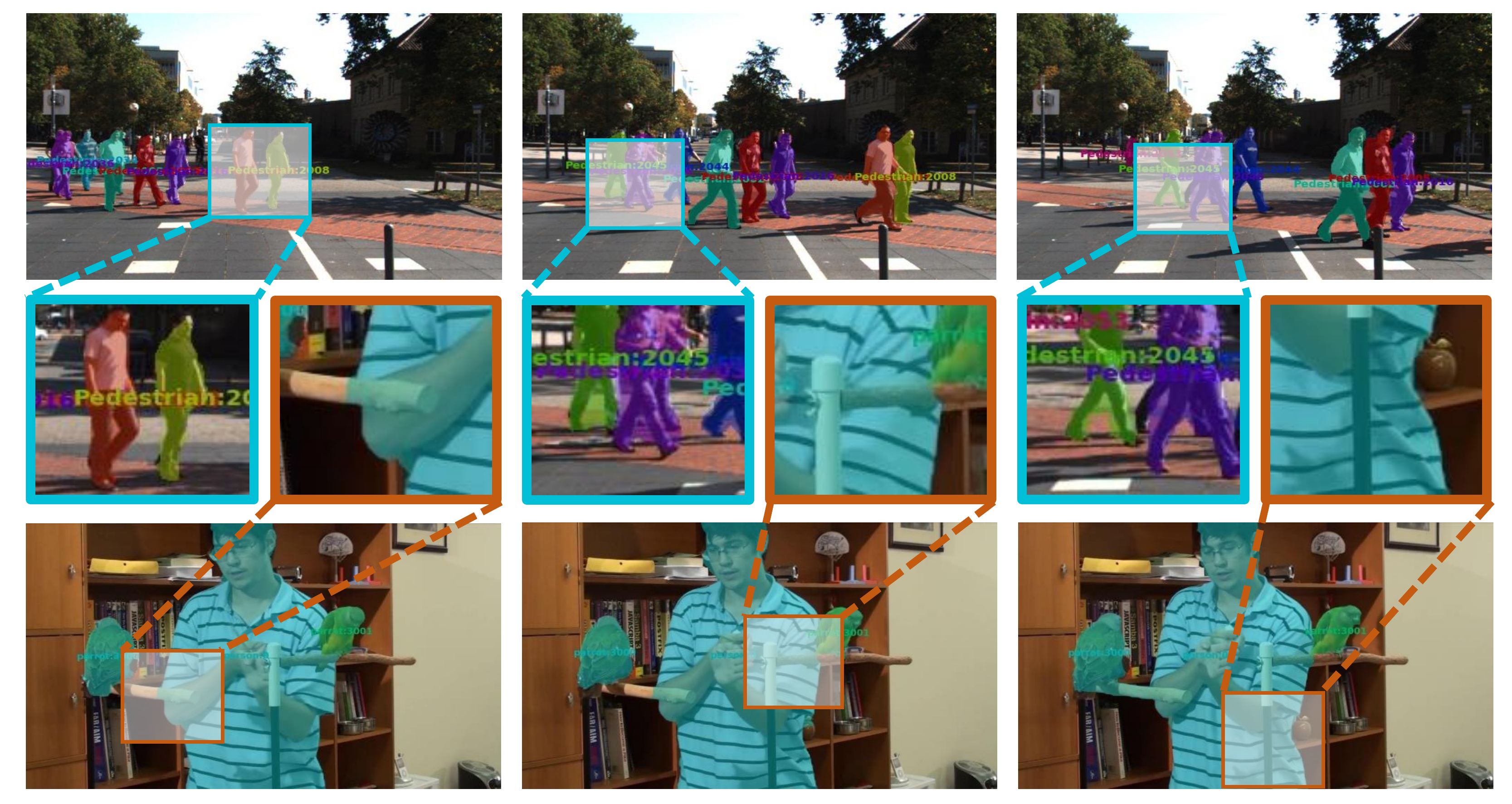}
    \caption{\textbf{Examples of weak predictions from STC-Seg.} The first row is from KITTI MOTS and the second row is from YT-VIS. }
    \label{fig:bad_example}
\end{figure}
The aforementioned results demonstrate the strong performance of STC-Seg in videos. We thus argue that
it is effective to
use the proposed pseudo-labels and puzzle solver to supervise the mask generation, especially for rigid objects (\eg, vehicles, boats, and planes). However, we encounter notable performance degradation for non-rigid objects (\eg, humans and animals) as the depth and flow estimation become less accurate under the circumstance, which compromises the corresponding pseudo-label generation for supervision.
For instances in Fig.~\ref{fig:bad_example}, there are large false positive regions between  pedestrian legs (the top row); our method fails to segment objects in front of the man (the bottom row). The above weak predictions are primarily caused by noisy pseudo-labels incurred by inaccurate depth and flow estimation.

\begin{figure*}[t]
    \centering
    \includegraphics[width=0.7\linewidth]{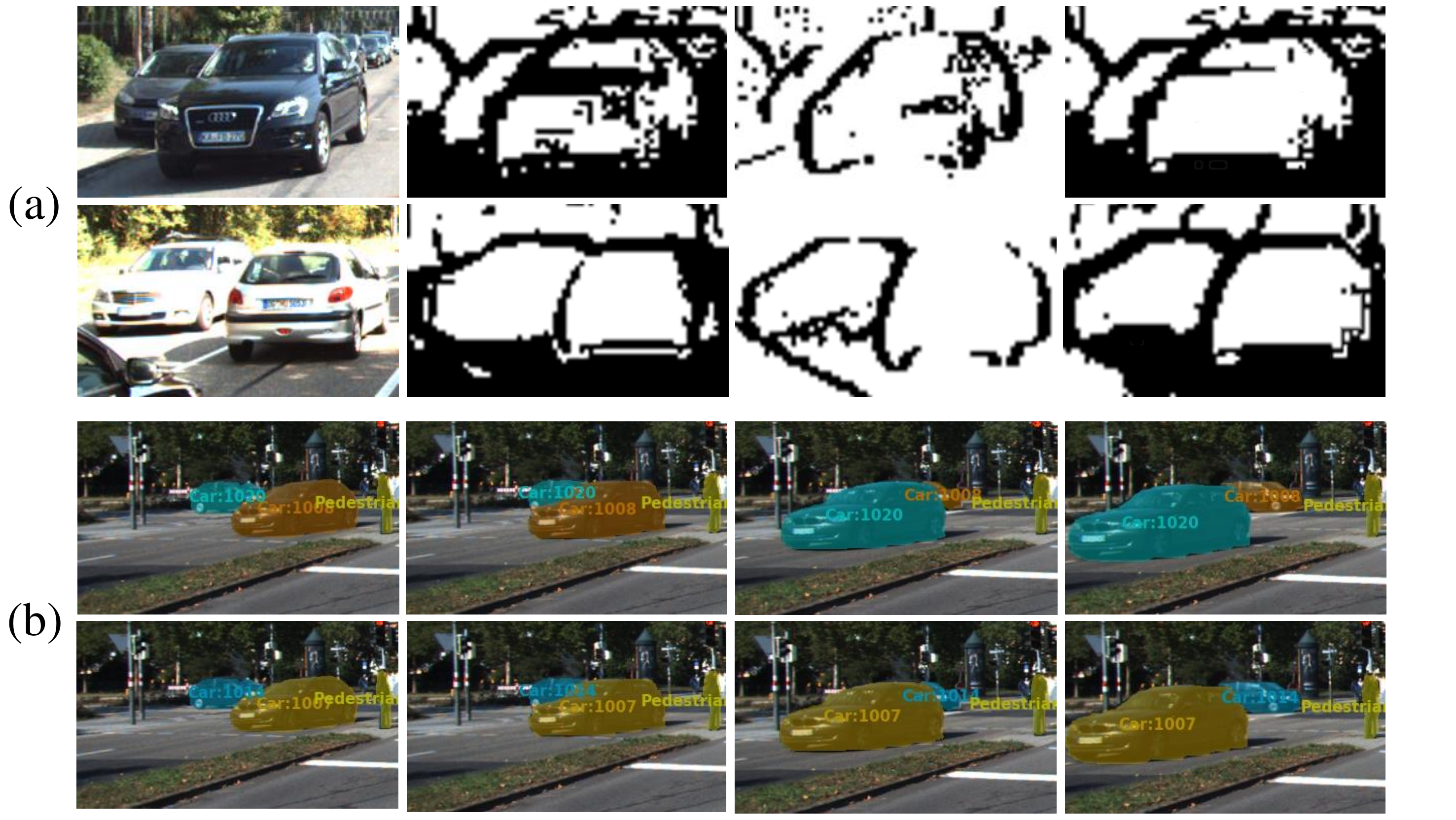}
    \caption{\textbf{Ablation study results}: a). The pseudo-labels generated by depth signals ($2^{nd}$ column), optical flow signals ($3^{rd}$ column), and combination of both ($4^{th}$ column); b). The same mask color indicates the same instance. The first row results are from CP \cite{Zhou2020TrackingOA}, which often encounters the issue of ID switching. The second row results are from ours, which is robust to object appearance changes.
    }
    \label{fig:ablation_study}
    \vspace{-8pt}
\end{figure*}

\subsection{Ablation Study}\label{Understanding_STC_Seg}
In this section, we investigate the effectiveness of each component in STC-Seg by conducting ablation experiments on KITTI MOTS. For the assessment of our supervision signals and loss terms, we focus on the improvement of mask generation and thus include the average precision (AP) in evaluation. To assess our tracking, we use HOTA, MOTSA, and MOTSP from MOTS~\cite{luiten2021hota}.

\begin{table}[tbh]
\caption{The results of \textbf{using different supervision signals} from KITTI MOTS. C and P denote \textit{car} and \textit{pedestrian} respectively.}
    \label{tab:different_supervision_signals}
    \centering
    
    \begin{tabular}{c|cc|cc|cc}
         \hlineB{2.5}
	     \rowcolor{mygray} 
         & \multicolumn{2}{c|}{AP} & \multicolumn{2}{c|}{HOTA} & \multicolumn{2}{c}{MOTSA} \\
         \rowcolor{mygray} 
         \multirow{-2}{*}{Signal} & C & P & C & P & C & P \\
         
         \hline
         \midrule
         
         Depth & 55.6 & 37.5 & 57.7 & 45.0 & 82.1 & 66.2 \\
         Flow & 55.0 & 37.9 & 56.8 & 46.2 & 80.7 & 66.9 \\
         Depth+Flow & \textbf{56.1} & \textbf{38.2} & \textbf{59.6} & \textbf{47.5} & \textbf{83.3} & \textbf{68.3} \\
         \bottomrule
    \end{tabular}
    
\end{table}
\begin{table}[tbh]
\caption{The results of \textbf{using different loss terms} from KITTI MOTS. C and P denote \textit{car} and \textit{pedestrian} respectively.
    }
    \label{tab:different_loss_term}
    \centering
    
    \begin{tabular}{c|cc|cc|cc|cc}
         \hlineB{2.5}
	     \rowcolor{mygray} 
         \multicolumn{3}{c|}{AP} & \multicolumn{2}{c|}{AP} & \multicolumn{2}{c|}{HOTA} & \multicolumn{2}{c}{MOTSA} \\
         \rowcolor{mygray} 
         $L_{bx}$ & $L_{bce}$ & $L_{bd}$ & C & P & C & P & C & P \\
         \hline
         \midrule
         \multirow{3}{*}{\rotatebox{90}{$L_{dice}$}} 
         & $\times$ & $\times$ & 53.3 & 34.5 & 53.9 & 40.3 & 78.1 & 61.7 \\
         & \checkmark & $\times$ & 53.8 & 35.0 & 54.8 & 42.2 & 79.2 & 63.0 \\
         & $\times$ & \checkmark & 54.7 & 36.9 & 56.7 & 44.1 & 81.6  & 65.4 \\
         \midrule
         \multirow{3}{*}{\rotatebox{90}{$L_{dice}'$}} 
         & $\times$ & $\times$ & 53.6 & 34.9 & 54.6 & 41.9 & 79.0 & 62.5 \\ 
         & \checkmark & $\times$ & 54.2 & 36.4 & 55.4 & 43.2 & 79.7 & 64.6 \\
         & $\times$ & \checkmark  & \textbf{56.1} & \textbf{38.2} & \textbf{59.6} & \textbf{47.5} & \textbf{83.3} & \textbf{68.3} \\
         \bottomrule
    \end{tabular}
    
\end{table}

\begin{table}[tbh]
\caption{
    The impact of \textbf{different mini network depth} in KITTI MOTS. C and P denote \textit{car} and \textit{pedestrian} respectively.}
    
    \label{tab:different_mini_network_depth}
    \centering
    
    \begin{tabular}{c|cc|cc|cc}
         \hlineB{2.5}
	     \rowcolor{mygray} 
         & \multicolumn{2}{c|}{HOTA} & \multicolumn{2}{c|}{MOTSA}  & \multicolumn{2}{c}{MOTSP} \\
         \rowcolor{mygray} 
         \multirow{-2}{*}{Depth} & C & P & C & P & C & P \\
         \hline
         \midrule
         1  & 57.4 & 46.0   & 82.3   & 67.4  & 83.3  & 73.8  \\
         2  & 58.8  & 46.6   & 82.8  & 67.9  & 84.2  & 74.7  \\
         3  & \textbf{59.6} & \textbf{47.5} & \textbf{83.3} & \textbf{68.3} & \textbf{85.1} & \textbf{75.8} \\
         4  & 59.2   & 47.1  & 83.1  & 68.2  & 84.9  & 75.3  \\
         \bottomrule
    \end{tabular}
    
\end{table}

\begin{table}[tbh]
\caption{
    The impact of \textbf{different mini network dilation rate} in KITTI MOTS. The depth of the mini network is fixed to 3. C and P denote \textit{car} and \textit{pedestrian} respectively.}
    
    \label{tab:different_mini_network_dilation}
    \centering
    
    \begin{tabular}{c|cc|cc|cc}
         \hlineB{2.5}
	     \rowcolor{mygray} 
         & \multicolumn{2}{c|}{HOTA} & \multicolumn{2}{c|}{MOTSA}  & \multicolumn{2}{c}{MOTSP} \\
         \rowcolor{mygray} 
         \multirow{-2}{*}{$\lambda$} & C & P & C & P & C & P \\
         \hline
         \midrule
         1  & 56.7   & 45.4  & 81.9  & 66.8  & 83.0  & 73.1  \\
         2  & \textbf{59.6} & \textbf{47.5} & \textbf{83.3} & \textbf{68.3} & \textbf{85.1} & \textbf{75.8} \\
         3  & 57.9  & 46.7  & 82.5  & 67.2  & 84.3  & 74.4  \\
         \bottomrule
    \end{tabular}
    
\end{table}

\begin{table}[tbh]
\caption{
    The impact of \textbf{different mini network filter factors} in KITTI MOTS. The results are obtained by HOTA on Car and Pedestrian Category respectively.}
    
    \label{tab:different_mini_network_filter}
    \centering
    
    \begin{tabular}{c|cccc}
         \hlineB{2.5}
	     \rowcolor{mygray} 
         \diagbox{$\phi^s$}{$\phi^t$} & 0.2 & 0.3 & 0.4 & 0.5 \\
         \hline
         \midrule
           0.2 & 38.2 / 32.8 & 43.6 / 37.4 & \underline{54.1} / 43.2 & 48.7 / 36.5 \\
           0.3 & 42.6 / 37.7 & 49.3 / \textbf{47.9} & \textbf{59.6} / \underline{47.5} & 52.6 / 40.1 \\
           0.4 & 39.0 / 35.4 & 46.8 / 41.1 & 50.6 / 40.4 & 41.9 / 35.3 \\

         \bottomrule
    \end{tabular}
    
\end{table}

\begin{table}[tbh]
\caption{
    The results of \textbf{using different tracking strategies} from KITTI MOTS.  CP, DP, and SD are short for center point, diagonal points, and spatio-temporal discrepancy respectively.
    }
    \label{tab:different_tracking_strategy}
    \centering
    
    \begin{tabular}{c|cc|cc|cc}
         \hlineB{2.5}
	     \rowcolor{mygray} 
         & \multicolumn{2}{c|}{HOTA} & \multicolumn{2}{c|}{MOTSA}  & \multicolumn{2}{c}{MOTSP} \\
         \rowcolor{mygray} 
         \multirow{-2}{*}{Tracking by} & C & P & C & P & C & P \\
         \hline
         \midrule
         CP  & 58.9 & 47.1 & 83.1 & 68.0 & 83.9 & 74.5 \\
         DP  & 59.3 & 47.3 & 83.2 & 68.2 & 84.7 & 74.9 \\ 
         DP+SD  & \textbf{59.6} & \textbf{47.5} & \textbf{83.3} & \textbf{68.3} & \textbf{85.1} & \textbf{75.8} \\
         \bottomrule
    \end{tabular}
    
\end{table}

\begin{table}[tbh]
\caption{The comparison results on YT-VIS validation set. * and $^\dagger$ indicate the use of the \textbf{ground truth} and \textbf{pseudo labels} respectively during training. All methods use ResNet-101 with FPN.
    }
    \label{i2v}
    \centering
    
    \begin{tabular}{c|c|c|c}
         \hlineB{2.5}
	     \rowcolor{mygray} 
         Methods & mAP & AP@0.75 & AR@10 \\
         \hline
         \midrule
           YOLACT* & 29.7&32.1 &36.5   \\
           BlendMask*  &32.0& 34.1&39.7  \\
           HTC*        &35.3 &36.9  &40.8   \\
                   \midrule
                   YOLACT$^\dagger$ &28.9&31.2 &35.8   \\
           BlendMask$^\dagger$
           &31.3& 33.3&38.8  \\
           HTC$^\dagger$ &34.5 &36.3  &40.0   \\

         \bottomrule
    \end{tabular}
    
\end{table}

\subsubsection{Supervision Signals} 
We show the impact of progressively integrating the depth and flow signals for the pseudo-label generation. As shown in Table~\ref{tab:different_supervision_signals}, compared to optical flow, depth has a better performance for car class to produce pseudo-labels when being used alone, while optical flow has a better performance for pedestrian class. In contrast, by leveraging both depth and flow, we develop complementary representations that retain richer and more accurate details of the instance boundary for pseudo-label generation (see Fig.~\ref{fig:ablation_study}a). Therefore, combining the two signals together enables our model to achieve the best performance over the baselines that use them separately.

\subsubsection{Loss Terms} We first only use our box term  $L_{bx}(L_{dice})$ without the position penalty to supervise the mask generation as our baseline, followed by the variants supervised by different loss combinations
(see Table~\ref{tab:different_loss_term}). We achieve immediate improvements of 2.8\% (car) and 3.7\% (pedestrian) on AP for the model trained only by $L_{bce}$+$L_{bx}(L_{dice})$ over the baseline. While using BCE loss $L_{bce}$ and our $L_{bx}(L_{dice}')$ for supervision, we can obtain further performance gain over the models trained by $L_{bce}$+$L_{bx}(L_{dice})$. The best results come from the model trained by our puzzle loss $L_{bd}$+$L_{bx}(L_{dice}')$, whose margins over the second best results ($L_{bd}$+$L_{bx}(L_{dice})$) by 1.4\% (car) and 1.3\% (pedestrian) on AP. The above results confirm our assumption for our puzzle loss design that  
the proposed box term and boundary term can work collaboratively to generate a high-quality instance mask.

\subsubsection{Mini Network Architecture}
We also evaluate the impact of using different configurations for the mini network. Specifically, we vary the mini network depth (number of
layers) from the list of $\{1, 2, 3, 4\}$ with the fixed dilation rate $\lambda$ of 2 and dilation convolution kernel size 3. We also vary the dilation rate $\lambda$ of the mini network from $\{1, 2, 3\}$, and use the grad search to determine the filter factors $\phi^s, \phi^t$. The results are shown in Table \ref{tab:different_mini_network_depth}, Table \ref{tab:different_mini_network_dilation} and Table \ref{tab:different_mini_network_filter} respectively. Those results show that a reasonable mini network configuration can account for better supervision, where the mini network includes 3 layers of dilated convolutions with a dilation rate of 2 and a kernel size of 3. To achieve better performances, the filter factors $\phi^s, \phi^t$ are set to be 0.3 and 0.4 respectively.

\subsubsection{Tracking Strategy}
We finally evaluate the impact of using different elements for tracking (see Table~\ref{tab:different_tracking_strategy}). For CP, we use the state-of-the-art CenterTrack~\cite{Zhou2020TrackingOA}. For DP, we only use diagonal points in our tracking module. For DP+SD, it uses both diagonal points and spatio-temporal discrepancy. From the results we can see that DP provides immediate improvements in tracking over the baseline that uses CP. DP+SD further improves the tracking capacity compared to DP, demonstrating strong tracking robustness (see Fig.~\ref{fig:ablation_study}b).
These results suggest that each element (\ie \ DP and SD) individually contributes towards improving the tracking performance.

\subsection{Extending Instance Segmentation to Videos}\label{toVideo}
In this section, we evaluate the flexibility and generalization of the proposed STC-Seg framework.
In particular, we leverage our STC-Seg framework (\ie \ pseudo-label generation, puzzle solver, and tracking module) to extend image instance segmentation methods to the video task. We select three widely recognized instance segmentation methods, YOLACT~\cite{bolya2019yolact},   BlendMask~\cite{chen2020blendmask} and HTC \cite{chen2019hybrid}), and integrate with our STC-Seg framework. The results of two set of experiments, \ie \ training with ground truth labels and pseudo labels, on YT-VIS are shown in Table~\ref{i2v}.
Each of the selected methods is crafted with our tracking module and uses the same implementation as discussed in Section \ref{Experiments}. 
It can be seen that methods trained using the proposed pseudo-labels achieve comparable results with the models trained on ground truth labels. This observation is consistent among all three selected methods, which demonstrates that our STC-Seg framework can flexibly extend image instance segmentation methods to operate on video tasks. 

\begin{table}[tbh]
\caption{\textbf{Results of using ground truths or not in spatio-temporal signals generation} when training our STC-Seg on KITTI MOTS. ``$\times$" denotes signal is obtained from the predicted depth or flow, while ``\checkmark" denotes signal is obtained from their ground truth. C and P denote \textit{car} and \textit{pedestrian} respectively.
    }
\label{tab:using_gt}
\vspace{10pt}
    \centering
    
    \begin{tabular}{cc|cc|cc|cc}
         \hlineB{2.5}
	     \rowcolor{mygray}
         \multicolumn{2}{c|}{Signal GT} & \multicolumn{2}{c|}{HOTA} & \multicolumn{2}{c|}{MOTSA} & \multicolumn{2}{c}{MOTSP} \\
         \rowcolor{mygray}
         Depth & Flow & C & P & C & P & C & P \\
         \hline
         \midrule
         $\times$ & $\times$ & 59.6 & 47.5 & 83.3 & 68.3 & 85.1 & 75.8  \\ 
         \checkmark & $\times$ & 63.1 & 48.3 & 89.9 & 69.0 & 86.6 & 75.9 \\
         $\times$ & \checkmark & 61.4 & 49.2 & 87.5 & 69.8 & 86.4  & 76.1 \\
         \checkmark  & \checkmark  & \textbf{64.2} & \textbf{51.1} & \textbf{92.7} & \textbf{70.5} & \textbf{86.9} & \textbf{76.2} \\
         \bottomrule
    \end{tabular}
    
\end{table}

\subsection{Results Using Ground Truth Depth and Flow}
Since depth estimation and optical flow are critical factors to generate our pseudo-label, we also directly employ the ground truth depth and flow for the pseudo-label generation in training to investigate the performance gap between using the predicted spatio-temporal signals and ground truths. Table \ref{tab:using_gt} demonstrates the results on KITTI MOTS. We can see that using depth and flow ground truths can further improve the performance. Thus, we argue that with strong depth and flow predictions, our method can achieve further performance gain.

\section{Conclusion and Limitation}
Instance segmentation in videos is an important research problem, which has been applied in a wide range of vision applications.
In this study, we propose a weakly supervised learning method for instance segmentation in videos with a spatio-temporal collaboration framework, titled STC-Seg. In particular, we introduce a weakly supervised training strategy which successfully combines unsupervised spatio-temporal collaboration and weakly supervised signals, helping networks to jointly achieve completeness and adequacy for instance segmentation in videos without pixel-wised labels. STC-Seg works in a plug-and-play manner and can be nested in any segmentation network method. Extensive experimental results indicate that STC-Seg is  competitive  with  the concurrent methods and outperforms fully supervised MaskTrack R-CNN and TrackR-CNN.
Albeit achieving strong performance, our method requires box labels to operate training which limits its applicability to new tasks without any prior knowledge. This challenge remains open for our future research endeavors. 
There are several ongoing investigations. For example, we are exploring unsupervised or weakly supervised object detection methods to obtain box labels. These predicted box labels can then be used to predict instance segmentation.






\bibliographystyle{IEEEtran}
\bibliography{egbib.bib}

\end{document}